\documentclass{article}

\usepackage[preprint]{neurips_2026}

\usepackage{graphicx}
\usepackage{amsmath,amssymb}
\usepackage{hyperref}
\usepackage{xcolor}
\usepackage{booktabs}
\usepackage{float}
\usepackage{caption}
\usepackage{subcaption}
\usepackage{algorithm}
\usepackage{algorithmic}
\usepackage{wrapfig}
\usepackage{multirow}
\usepackage{pifont}

\hypersetup{
    colorlinks=true,
    linkcolor=blue!70!black,
    citecolor=blue!70!black,
    urlcolor=blue!70!black
}

\title{Central Dogma Transformer III:\\
Interpretable AI Across DNA, RNA, and Protein}

\author{
  Nobuyuki Ota \\
  Independent Researcher\\
  Burlingame, CA, USA \\
  \texttt{nobuyuki.ohta@gmail.com}
}

\begin{document}

\maketitle

\begin{abstract}
Biological AI models increasingly predict complex cellular responses, yet their learned representations remain disconnected from the molecular processes they aim to capture---an interpretability gap that limits scientific insight. Closing this gap requires architectures that mirror the cell's own logic---the central dogma's information flow from DNA to RNA to protein---so that each computational step corresponds to a verifiable biological process. We present CDT-III, which extends mechanism-oriented AI across the full central dogma: DNA, RNA, and protein. Its two-stage architecture mirrors the spatial compartmentalization of the cell: a Virtual Cell Embedder for the nucleus (VCE-N) models transcription and one for the cytosol (VCE-C) models translation. By following the cell's molecular pathway, CDT-III produces interpretable attention maps at every layer and jointly predicts mRNA and surface protein changes from CRISPRi perturbations. On five held-out genes, CDT-III achieves per-gene RNA $r = 0.843$ and protein $r = 0.969$. Adding protein prediction improves RNA performance over the RNA-only model ($r = 0.804 \to 0.843$), demonstrating that downstream tasks can regularize upstream representations. Protein supervision also sharpens DNA-level interpretability, increasing CTCF enrichment by 30\%. Applied to in silico CD52 knockdown---approximating Alemtuzumab's mechanism of action---the model predicts 29/29 protein changes in the correct direction and rediscovers 5 of 7 known clinical side effects without clinical data. Moreover, gradient-based side effect profiling requires only unperturbed baseline data as input ($r = 0.939$ vs.\ perturbation-based analysis), enabling screening of all 2,361 modeled genes without new experiments. This demonstrates that mechanism-oriented AI, designed to reveal rather than merely predict, can generate clinically actionable insights from perturbation data alone.
\end{abstract}

\section{Introduction}

The interpretability gap in biological AI---the disconnect between a model's predictions and the molecular mechanisms producing them---persists in part because existing architectures do not reflect the cell's own information flow~\citep{novakovsky2023,eraslan2019}. The central dogma~\citep{crick1970} routes information from DNA through RNA to protein, yet existing perturbation models stop at mRNA~\citep{roohani2024gears,lotfollahi2023cpa}, missing protein-level phenotypes entirely; mRNA abundance explains only 40--80\% of protein variance~\citep{gygi1999,vogel2012}. This architectural mismatch carries direct consequences for drug development, where mechanistic understanding at the protein level determines whether a side effect can be anticipated or only observed after the fact.

Despite this gap, substantial progress has been made in perturbation prediction~\citep{roohani2024gears,lotfollahi2023cpa}, single-cell foundation models~\citep{cui2024scgpt,theodoris2023}, and multi-modal biological AI~\citep{chen2024isoformer}. CDT-II~\citep{ota2026cdtii} demonstrated that mirroring the central dogma in model architecture yields attention maps directly interpretable as regulatory structure. However, no existing model connects all three layers of the central dogma---DNA, RNA, and protein---in an interpretable, end-to-end framework for perturbation response prediction.

We present CDT-III, which extends mechanism-oriented AI to the complete central dogma. The architecture comprises two stages---VCE-N (nuclear) and VCE-C (cytosolic)---that separate transcription and translation into distinct computational modules, each with attention mechanisms corresponding to specific biological processes. This biological inductive bias is not optional: a naive single-stage fusion fails to maintain RNA prediction quality (Section~\ref{sec:ablation}), confirming that architectural design must respect the spatial and functional organization of the cell.

CDT-III yields three findings with implications beyond biology:

\begin{enumerate}
    \item \textbf{Multi-task regularization}: Adding protein prediction improves RNA prediction, demonstrating that downstream tasks can regularize upstream representations.
    \item \textbf{Downstream supervision improves upstream interpretability}: Protein supervision sharpens DNA-level regulatory element detection by 30\%, showing that learning to predict distal phenotypes improves proximal understanding.
    \item \textbf{In silico pharmacology}: Applied to CD52 knockdown (approximating Alemtuzumab), CDT-III rediscovers known clinical side effects and generates novel testable hypotheses from perturbation data alone, without access to any clinical information.
\end{enumerate}

Together, these results suggest that AI architectures aligned with biological structure can simultaneously achieve prediction, interpretation, and clinical translation.

\section{Related Work}

\textbf{Perturbation prediction and foundation models.}
GEARS~\citep{roohani2024gears} and CPA~\citep{lotfollahi2023cpa} predict expression changes from unseen perturbations but are limited to RNA. Foundation models (scGPT~\citep{cui2024scgpt}, Geneformer~\citep{theodoris2023}, Enformer~\citep{avsec2021enformer}, Borzoi~\citep{linder2025borzoi}, AlphaGenome~\citep{avsec2026alphagenome}) provide powerful representations, though recent benchmarks show deep learning does not yet consistently outperform simple baselines~\citep{hetzel2025baselines,piran2024perteval}. STRAND~\citep{bunne2026strand} conditions perturbation prediction on DNA sequence for zero-shot inference but does not model protein responses. CDT-III differs in jointly predicting RNA and protein changes with interpretable internal representations.

\textbf{Multi-modal biological AI.}
IsoFormer~\citep{chen2024isoformer} integrates DNA, RNA, and protein for isoform prediction. LucaOne~\citep{he2025lucaone} provides a unified DNA/RNA/protein foundation model, and CAPTAIN~\citep{ji2025captain} pre-trains on 4M+ co-assayed RNA and protein cells, but neither addresses perturbation prediction or interpretability. CITE-seq~\citep{stoeckius2017} and DSB normalization~\citep{muleisenburger2020} enable single-cell protein quantification, and perturbation datasets are increasingly harmonized~\citep{peidli2024scperturb,heumos2025pertpy}. Despite these resources, no existing model jointly predicts RNA and protein perturbation responses in an interpretable framework.

\textbf{Drug safety and interpretability.}
Drug side effect prediction typically relies on chemical structure~\citep{zitnik2018decagon} or network proximity~\citep{guney2016}, requiring known drug--target interactions. CDT-III derives predictions from learned gradients without clinical data. Chen et al.~\citep{chen2024iml} emphasize rigorous interpretability practices, warning against single-method reliance and disconnection from biological validation. CDT-II~\citep{ota2026cdtii} established mechanism-oriented AI with attention maps validated against external data, addressing the attention-as-explanation debate~\citep{jain2019attention,wiegreffe2019attention}. CDT-III extends this to clinical application, validating attention against CTCF ChIP-seq and Hi-C data.

\section{Method}

\subsection{Problem Formulation}

Given a CRISPRi perturbation at genomic locus $\ell$ and a cell with RNA expression $\mathbf{x}_{\text{RNA}} \in \mathbb{R}^{2361}$ and surface protein levels $\mathbf{x}_{\text{Prot}} \in \mathbb{R}^{189}$ (DSB-normalized), CDT-III predicts perturbation-induced changes at both molecular levels:
\begin{equation}
f(\mathbf{D}_\ell,\; \mathbf{x}_{\text{RNA}},\; \mathbf{x}_{\text{Prot}}) = (\hat{\mathbf{y}}_{\text{RNA}},\; \hat{\mathbf{y}}_{\text{Prot}})
\end{equation}
where $\mathbf{D}_\ell \in \mathbb{R}^{896 \times 3072}$ are pre-computed Enformer~\citep{avsec2021enformer} embeddings spanning $\sim$115\,kb around locus $\ell$, $\hat{\mathbf{y}}_{\text{RNA}} \in \mathbb{R}^{2361}$ are predicted RNA log2 fold changes, and $\hat{\mathbf{y}}_{\text{Prot}} \in \mathbb{R}^{189}$ are predicted protein expression differences relative to unperturbed cells.

\subsection{Architecture: Two-Stage Virtual Cell Embedder}

\begin{figure}[t]
\centering
\includegraphics[width=0.75\textwidth]{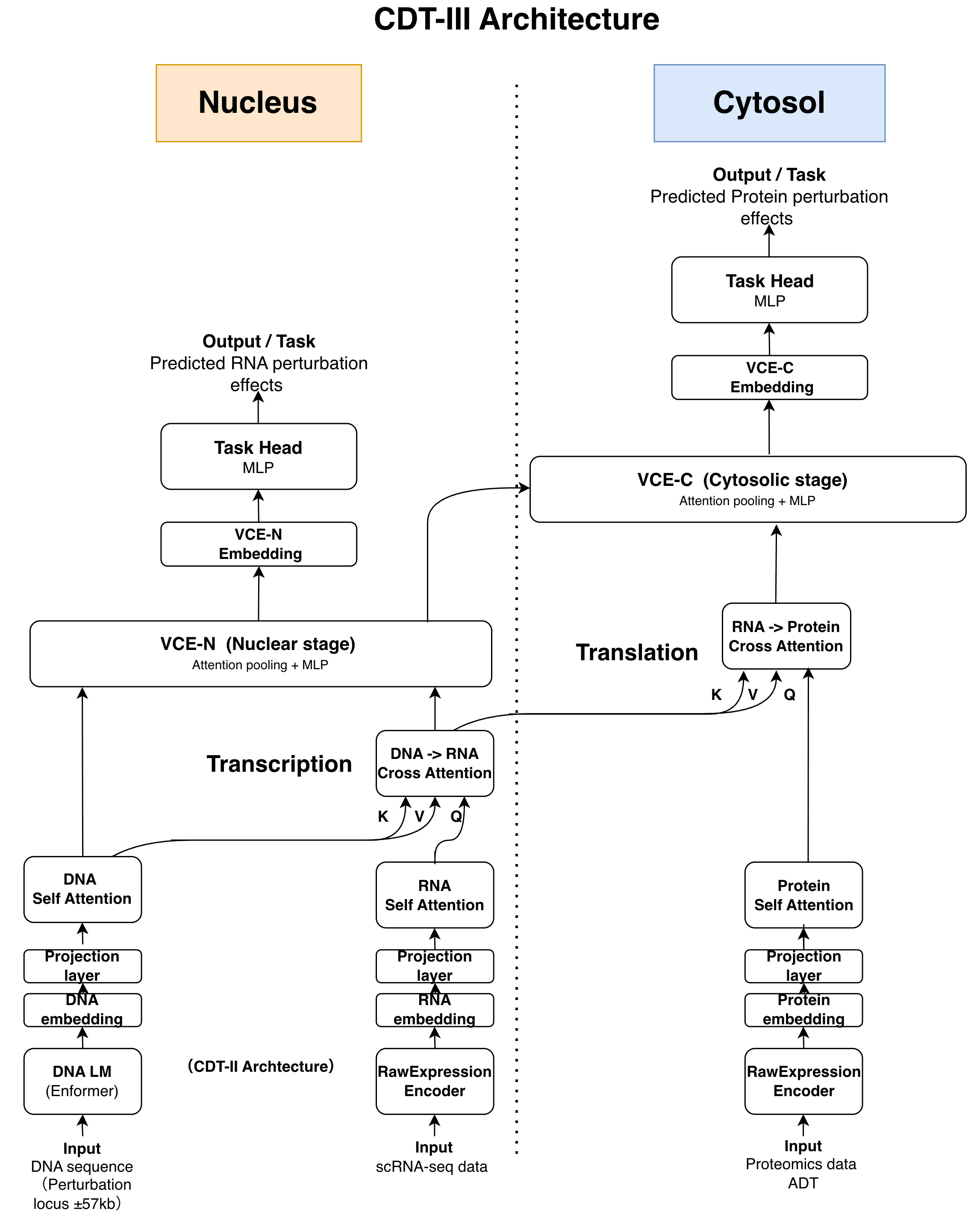}
\caption{\textbf{CDT-III two-stage Virtual Cell Embedder architecture.} VCE-N (nuclear stage, identical to CDT-II) processes DNA Enformer embeddings and per-cell RNA expression through self-attention and cross-attention, modeling transcription to produce a cell-level RNA embedding. VCE-C (cytosolic stage) takes this RNA embedding and protein expression, modeling translation to produce a protein embedding. Each stage has an independent task head.}
\label{fig:architecture}
\end{figure}

CDT-III models the complete information flow of the central dogma---from DNA to RNA to protein---as a single, end-to-end differentiable architecture (Figure~\ref{fig:architecture}). Six attention mechanisms each correspond to a specific biological process: two for genomic structure (DNA self-attention), one each for gene co-regulation (RNA self-attention) and protein co-regulation (protein self-attention), and two cross-attention layers modeling transcriptional (DNA$\to$RNA) and translational (RNA$\to$Protein) control.

This information flow is organized into two stages---VCE-N (Nuclear) and VCE-C (Cytosolic)---that mirror the spatial compartmentalization of gene expression in eukaryotic cells: transcription in the nucleus, translation in the cytosol. The key design principle is that each stage preserves dimensional compatibility with its predecessor, enabling 100\% transfer of pre-trained CDT-II weights---a property that proves essential, as single-stage alternatives that break this compatibility fail to learn (Section~\ref{sec:ablation}).

\textbf{Stage 1: VCE-N (Nuclear --- Transcription).}
Architecturally identical to CDT-II~\citep{ota2026cdtii}: DNA embeddings projected to $d = 512$ with 2 self-attention layers, RNA encoded with 1 self-attention layer, cross-attention (RNA queries, DNA keys/values) for transcriptional control, and VCE pooling fusing both into $\mathbf{e}_{\text{RNA}} \in \mathbb{R}^{512}$. An RNA task head predicts $\hat{\mathbf{y}}_{\text{RNA}}$ (Appendix~\ref{app:architecture}).

\textbf{Stage 2: VCE-C (Cytosolic --- Translation).}
Protein expression $\mathbf{x}_{\text{Prot}} \in \mathbb{R}^{189}$ is encoded using the same architecture as the RNA encoder (identity embeddings + value projection), followed by 1 self-attention layer. RNA-to-protein cross-attention (protein queries, RNA sequence representations as keys/values) allows each protein to attend to the full gene expression context. A second VCE pools $\mathbf{e}_{\text{RNA}}$ (from VCE-N) and the attention-pooled protein representation into $\mathbf{e}_{\text{Prot}} \in \mathbb{R}^{512}$ via the same fusion architecture ($d \times 2 \to d \times 2 \to d$). A protein task head predicts $\hat{\mathbf{y}}_{\text{Prot}}$.

\textbf{Loss function.}
\begin{equation}
\mathcal{L} = \mathcal{L}_{\text{RNA}} + \lambda \cdot \mathcal{L}_{\text{Prot}}, \quad \lambda = 0.1
\end{equation}
where $\mathcal{L}_{\text{RNA}}$ is MSE over all 2,361 genes. All 189 proteins pass through the encoder, self-attention, and cross-attention layers, but $\mathcal{L}_{\text{Prot}}$ is computed only on 65 detectably expressed proteins (DSB mean $> 0.5$), masking non-expressed proteins from the loss to prevent the model from learning trivial zero-to-zero mappings. The value $\lambda = 0.1$ was selected to balance RNA and protein objectives (see ablation in Section~\ref{sec:ablation}). Total parameters: 30,987,766 ($\sim$21M from CDT-II + $\sim$10M new).

\subsection{Training: Two-Phase Strategy}

To prevent catastrophic forgetting of CDT-II's learned representations, training proceeds in two phases.
\textbf{Phase 1} (300 epochs, patience 30): VCE-N frozen with CDT-II weights; only VCE-C trained (lr $= 10^{-3}$), allowing the protein pathway to initialize before interacting with pre-trained components.
\textbf{Phase 2} (500 epochs, patience 50): all parameters unfrozen with differential learning rates (CDT-II: $10^{-5}$, new: $5 \times 10^{-5}$).
CDT-III fine-tuning completes in $\sim$3 hours on a single A100 GPU (Appendix~\ref{app:architecture}).

\subsection{In Silico Pharmacology Pipeline}
\label{sec:pipeline}

CDT-III enables two complementary approaches to drug safety assessment. For perturbations with matched RNA and protein measurements, \textit{direct prediction} simultaneously predicts RNA and protein changes, revealing which surface proteins are affected and in which direction. For novel perturbations not present in the training data, \textit{gradient analysis} ($\partial \hat{y}_{\text{Prot},j} / \partial x_{\text{RNA},g}$) quantifies how each surface protein $j$ responds to changes in gene $g$'s expression, yielding a protein-level side effect profile from the trained model's weights alone---without requiring new experimental data. Neither approach requires clinical or pharmacological data---predictions derive entirely from in vitro CRISPRi experiments and the model's learned multi-modal relationships. Note that CRISPRi knockdown approximates but does not recapitulate antibody drug action (see Limitations).

\section{Experimental Setup}
\label{sec:setup}

\textbf{Dataset.} We used the STING-seq v2 dataset~\citep{morris2023} (GEO: GSE171452), which jointly profiles scRNA-seq and 193 surface proteins (CITE-seq) in K562 cells subjected to CRISPRi perturbations. K562 is one of the most extensively characterized human cell lines (ENCODE, Roadmap Epigenomics, 4DN), providing the external validation data used throughout this study, and as a chronic myelogenous leukemia line, is directly relevant to hematological drug discovery. Multi-modal perturbation datasets combining CRISPRi, scRNA-seq, and surface protein measurements remain rare; K562 STING-seq is currently the only large-scale dataset with all three modalities. After guide assignment filtering, we retained 8,250 TSS-perturbed cells and 2,078 non-targeting control (NTC) cells. Of 193 ADTs, 4 isotype controls were removed, leaving 189 surface proteins quantified by DSB normalization~\citep{muleisenburger2020}, which uses isotype controls for robust background correction.

\textbf{Gene/protein sets.} 2,361 RNA genes were selected by cross-dataset reproducibility filtering (intersection of two independent CRISPRi screens~\citep{morris2023,gasperini2019}). Of 189 ADT proteins, 65 are detectably expressed (mean DSB $> 0.5$); protein loss and evaluation are restricted to these 65.

\textbf{Data splits.} Training: 6,213 TSS-perturbed cells from 23 perturbation targets. Validation: 2,037 cells from five perturbation targets held out entirely at the gene level---no perturbation data for these genes appears in training, ensuring that predictions reflect genuine generalization: \textit{GFI1B} (477 cells), \textit{TFRC} (847), \textit{TNFSF9} (349), \textit{CD44} (233), \textit{CD52} (131).

\textbf{Evaluation.} Prediction performance is measured by Pearson $r$ at two levels: per-cell (correlation across all genes within each cell) and per-gene mean (mean predicted vs.\ mean observed expression change for each held-out gene, averaged across genes). Interpretability is assessed by CTCF ChIP-seq enrichment (ENCODE K562) and Hi-C 3D contact ratios (4DN K562, 5 kb resolution). Clinical validation uses systematic comparison with published Alemtuzumab side effect profiles.

\section{Results}

\subsection{Architectural Design Matters: Single-Stage vs.\ Two-Stage}
\label{sec:ablation}

A naive approach to adding protein prediction would fuse all three modalities in a single VCE layer. We tested five such single-stage variants (Appendix~\ref{app:ablation}): all fail, with the best achieving per-cell RNA $r = 0.37$ vs.\ CDT-II's per-cell $r = 0.64$. The failure traces to modifying VCE fusion dimensions ($d \times 2 \to d \times 3$), which prevents transfer of CDT-II's pre-trained fusion weights---the critical bottleneck, as encoder and attention weights transfer successfully.

CDT-III's two-stage design resolves this by preserving VCE-N's dimensions exactly (100\% weight transfer) and adding VCE-C as a separate module.

\subsection{Prediction Performance}
\label{sec:prediction}

\begin{table}[t]
\centering
\caption{\textbf{Per-gene prediction performance.} Per-gene mean $r$ (pseudo-bulk) on five held-out validation genes. CDT-II values are from the same checkpoint used for CDT-III weight transfer. RNA per-gene mean $r$ improves from CDT-II ($0.804$) to CDT-III ($0.843$) despite adding a second task. Protein per-gene mean $r$ on 65 expressed proteins reaches $0.969$.}
\label{tab:prediction}
\begin{tabular}{lccccc}
\toprule
Gene & Cells & RNA $r$ (CDT-II) & RNA $r$ (CDT-III) & $\Delta$ & Protein $r$ (expr.) \\
\midrule
\textit{GFI1B} & 477 & 0.851 & 0.885 & +0.034 & 0.969 \\
\textit{TNFSF9} & 349 & 0.831 & 0.868 & +0.037 & 0.991 \\
\textit{TFRC} & 847 & 0.824 & 0.854 & +0.030 & 0.989 \\
\textit{CD44} & 233 & 0.795 & 0.858 & +0.063 & 0.935 \\
\textit{CD52} & 131 & 0.719 & 0.748 & +0.029 & 0.962 \\
\midrule
\textbf{Mean} & & \textbf{0.804} & \textbf{0.843} & \textbf{+0.039} & \textbf{0.969} \\
\bottomrule
\end{tabular}
\end{table}

\begin{figure}[t]
\centering
\includegraphics[width=\textwidth]{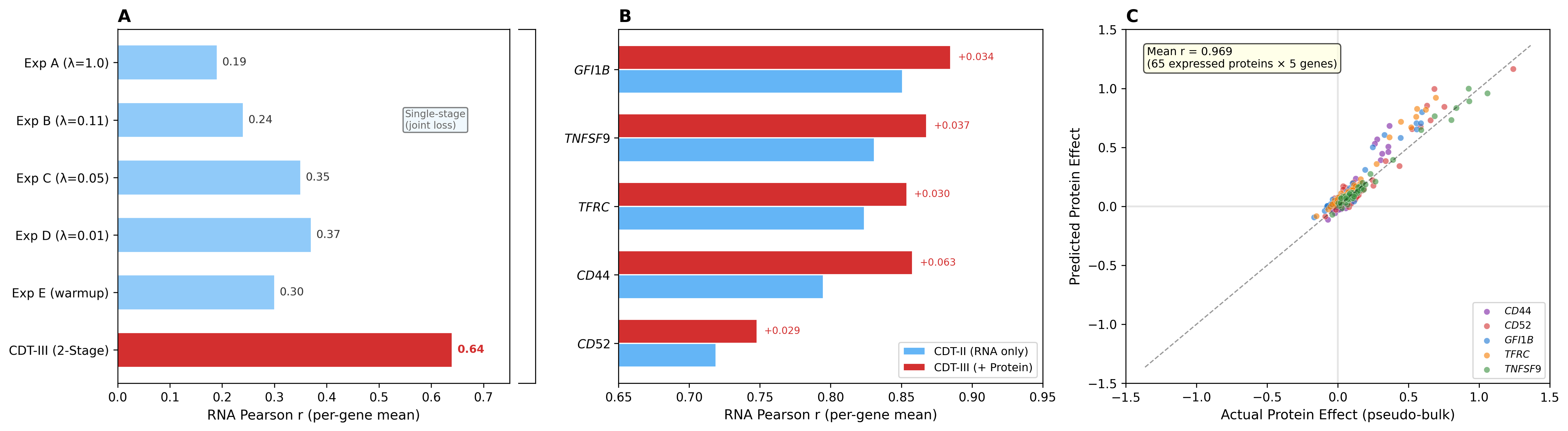}
\caption{\textbf{Prediction performance.} \textbf{(A)} Single-stage vs.\ two-stage VCE comparison (RNA $r$ by architecture). \textbf{(B)} Per-gene RNA correlation: CDT-II vs.\ CDT-III. \textbf{(C)} Per-gene protein predicted vs.\ actual (65 expressed proteins, 5 genes).}
\label{fig:prediction}
\end{figure}

\textbf{RNA prediction improves.} Adding protein prediction does not degrade RNA performance---it improves it (Table~\ref{tab:prediction}, Figure~\ref{fig:prediction}). The per-gene mean $r$ increases from $0.804$ to $0.843$ ($+4.9\%$). This multi-task regularization effect is consistent across all 5 genes, with improvement ranging from $+0.029$ (\textit{CD52}) to $+0.063$ (\textit{CD44}). The improvement indicates that protein loss gradients, backpropagating through VCE-C into VCE-N, provide additional supervision that sharpens the shared DNA and RNA representations.

\textbf{Protein prediction.} Of 189 surface proteins measured by CITE-seq, only 65 are detectably expressed in K562 cells. We evaluate exclusively on these 65 proteins to avoid inflating metrics with trivial zero-to-zero predictions (a pitfall that explains the apparent $r > 0.76$ of single-stage models in Table~\ref{tab:ablation}). On the 65 expressed proteins, per-gene protein $r$ averages $0.969$, with all 5 held-out genes exceeding $r = 0.93$ (Table~\ref{tab:prediction}). This indicates that CDT-III accurately captures which proteins change and in which direction following each perturbation.

Per-protein cell-level $r = 0.28$ is lower, consistent with known ADT measurement noise~\citep{stoeckius2017,muleisenburger2020}. The gap between cell-level and per-gene mean accuracy indicates that the model learns true regulatory relationships rather than fitting noise. Averaging across cells recovers the robust perturbation signal, positioning CDT-III for pseudo-bulk applications including drug safety assessment (Section~\ref{sec:cd52}).

\subsection{Protein Supervision Improves DNA Interpretability}

A surprising finding is that adding protein prediction improves CDT-III's ability to identify DNA regulatory elements, even though protein information does not directly enter the DNA pathway.

\textbf{CTCF enrichment.} DNA-to-RNA cross-attention in CDT-III shows $8.59\times$ enrichment at CTCF binding sites across 27 genes with CTCF sites (of 28 with ENCODE K562 ChIP-seq annotations; 1 gene has no CTCF sites), compared to $6.6\times$ for CDT-II ($+30\%$ improvement; Table~\ref{tab:ctcf}). All 27 genes with CTCF sites show $>2\times$ enrichment. Permutation testing ($n = 1{,}000$) yields $P < 0.001$.

\begin{table}[t]
\centering
\caption{\textbf{CTCF enrichment and Hi-C validation.} Protein supervision improves DNA-level interpretability.}
\label{tab:ctcf}
\begin{tabular}{lccc}
\toprule
Metric & CDT-II & CDT-III & Change \\
\midrule
Mean CTCF enrichment & $6.6\times$ & $8.59\times$ & $+30\%$ \\
Genes $>2\times$ enriched & --- & 27/27 (100\%)$^\dagger$ & --- \\
Permutation $P$ & $<0.001$ & $<0.001$ & --- \\
\midrule
Hi-C mean contact ratio & --- & $1.30\times$ & --- \\
Genes with ratio $>1$ & --- & 16/25 (64\%) & --- \\
Wilcoxon $P$ & --- & 0.020 & --- \\
\bottomrule
\end{tabular}
\begin{flushleft}
\footnotesize{$^\dagger$Of 28 genes with ENCODE K562 ChIP-seq annotations, 1 gene (MSR1) has no CTCF sites; enrichment is computed over the 27 genes with $\geq 1$ CTCF site.}
\end{flushleft}
\end{table}

\textbf{Mechanism.} Protein loss $\mathcal{L}_{\text{Prot}}$ backpropagates through VCE-C into the shared RNA embedding $\mathbf{e}_{\text{RNA}}$, which encodes both DNA and RNA information via DNA-to-RNA cross-attention. Because this embedding carries DNA-derived features, protein gradients flow further into VCE-N and the cross-attention weights, providing an additional signal that sharpens attention toward genomic positions that ultimately matter for protein-level phenotypes. CTCF binding sites---which demarcate topologically associating domain (TAD) boundaries and mediate 3D chromatin organization influencing both transcription and protein expression---are precisely such positions, explaining the $+30\%$ enrichment improvement.

\textbf{Attention reveals 3D chromatin structure at single-locus resolution.} Cross-attention not only enriches for CTCF sites in aggregate but resolves regulatory architecture at individual genomic loci. For CD55, the model's cross-attention profile across the $\pm$57\,kb window aligns precisely with CTCF ChIP-seq peaks: 3 of 4 CTCF binding sites fall within the top 10\% attention bins (Appendix~\ref{app:genome_tracks}). The single CTCF site ignored by the model lacks strong Hi-C contact with the promoter, indicating that the model distinguishes functionally relevant insulators from inactive ones. Hi-C data independently confirm this selectivity: across 25 genes with sufficient coverage (of 28; 3 excluded: \textit{B2M}, \textit{TFRC}, \textit{GFI1B}), high-attention bins show $1.30\times$ higher 3D contact frequencies with promoter regions than random bins ($P = 0.020$, Wilcoxon signed-rank; 4DN K562, 5\,kb resolution), with 16/25 genes showing ratios $> 1$. This genomic-position-level correspondence between learned attention, CTCF binding, and 3D chromatin contacts emerges entirely from perturbation prediction training---without any chromatin structure supervision.

\begin{figure}[t]
\centering
\begin{minipage}[t]{0.46\textwidth}
\vspace{0pt}
\includegraphics[width=\textwidth]{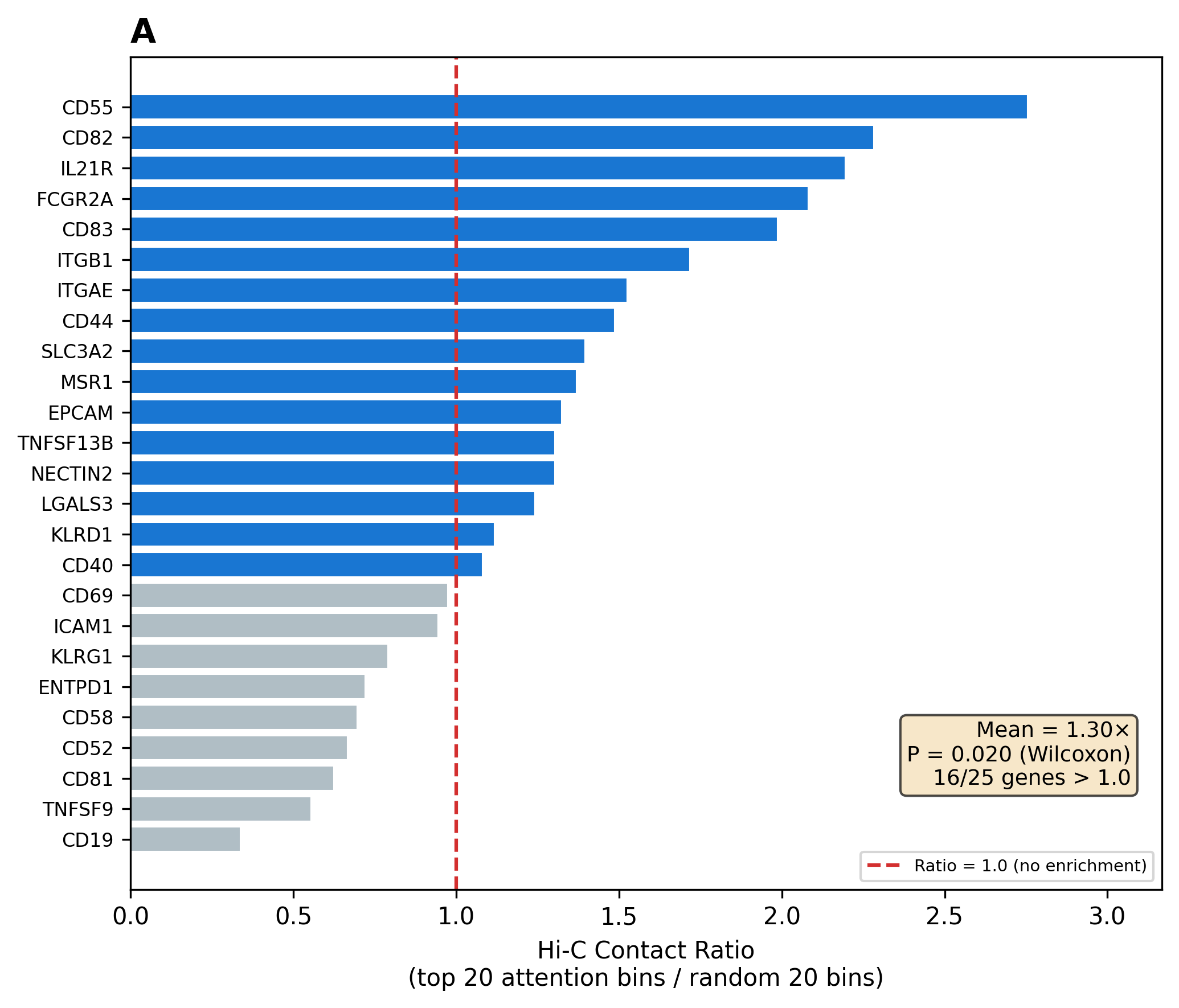}
\end{minipage}%
\hfill
\begin{minipage}[t]{0.36\textwidth}
\vspace{0pt}
\includegraphics[width=\textwidth]{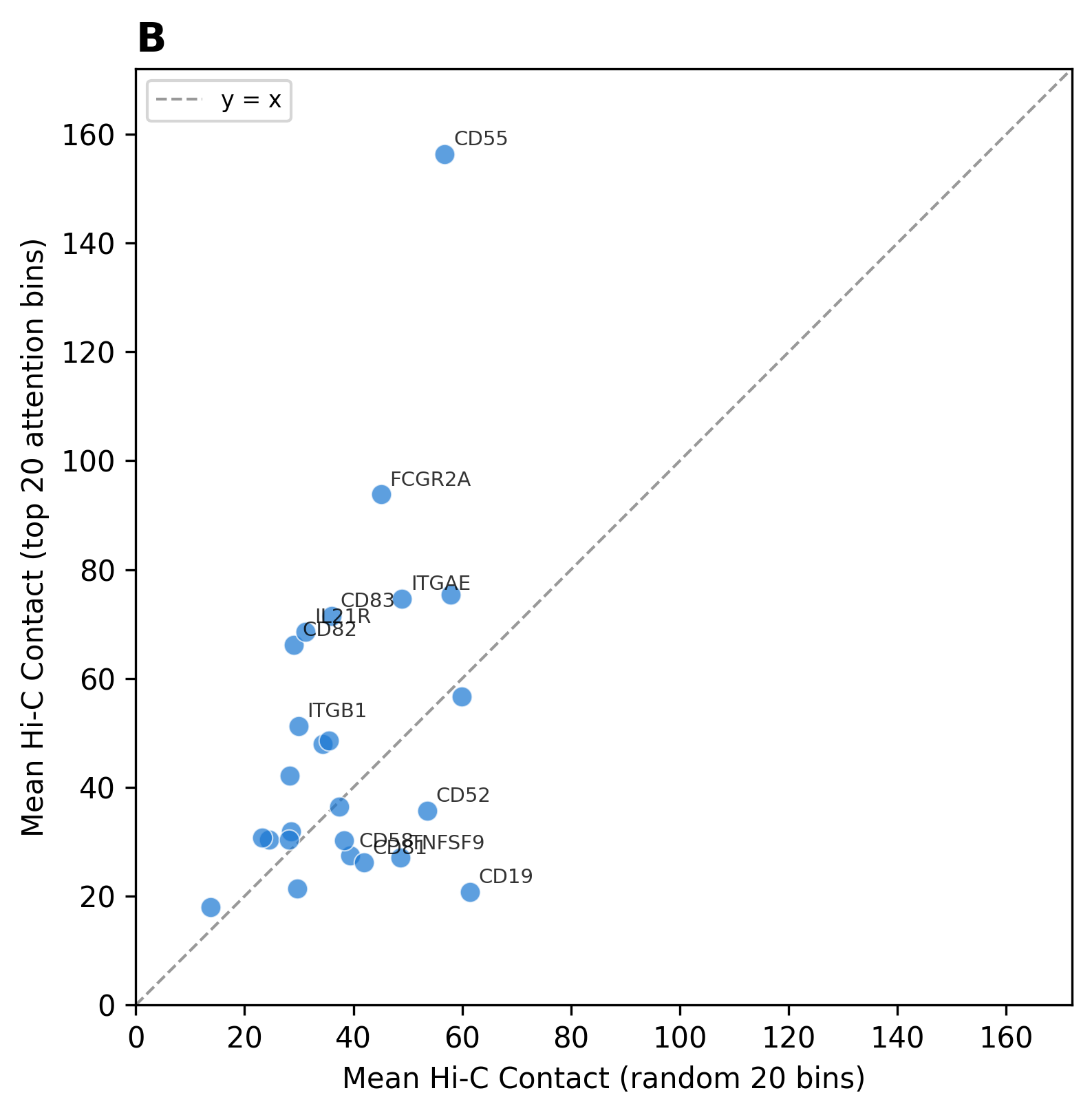}
\end{minipage}
\caption{\textbf{Protein supervision improves DNA interpretability.} \textbf{(A)} Hi-C contact ratio per gene (top 20 attention bins vs.\ random 20 bins; mean $= 1.30\times$, $P = 0.020$). \textbf{(B)} Attention-high vs.\ random bins scatter plot showing that high-attention genomic regions have stronger physical contact with promoters.}
\label{fig:ctcf}
\end{figure}
\subsection{In Silico Drug Safety: CD52/Alemtuzumab Case Study}
\label{sec:cd52}
\vspace{-2mm}
CD52 (CAMPATH-1 antigen) is the target of Alemtuzumab (Lemtrada\textsuperscript{\textregistered}), an FDA-approved monoclonal antibody for multiple sclerosis and chronic lymphocytic leukemia~\citep{ruck2022}. As discussed in Section~\ref{sec:pipeline}, CRISPRi knockdown of CD52 provides an in silico approximation of Alemtuzumab's mechanism of action. Critically, CD52 is one of five genes held out entirely from training---the model has never observed any CD52 perturbation data. We apply both direct prediction (131 CD52-perturbed cells from the held-out validation set) and gradient analysis (using only the non-targeting control mean expression and CD52's DNA embedding) to assess potential clinical consequences at the protein level.

\textbf{Prediction accuracy.} For CD52 knockdown, CDT-III achieves RNA per-gene $r = 0.748$ and protein $r = 0.962$ (65 expressed proteins). The self-knockdown prediction is highly accurate: mean predicted CD52 mRNA change $= -0.136$ vs.\ mean actual $= -0.172$ (per-cell $r = 0.952$ across 131 cells), confirming that CDT-III correctly predicts the direct effect of a perturbation on its own target.

\begin{figure}[t]
\centering
\begin{minipage}[c]{0.40\textwidth}
\includegraphics[width=\textwidth]{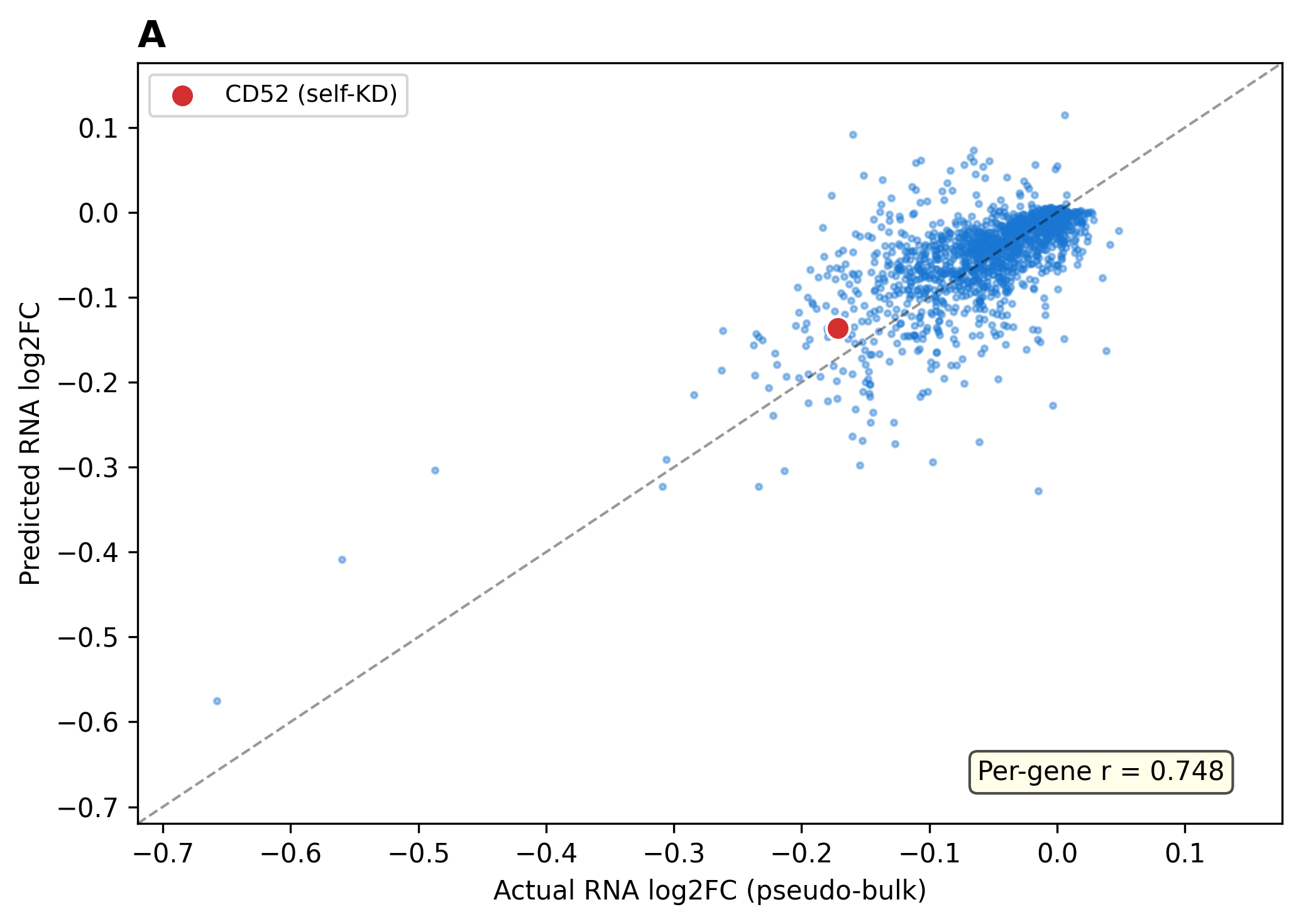}\\[1mm]
\includegraphics[width=\textwidth,height=3.8cm,keepaspectratio]{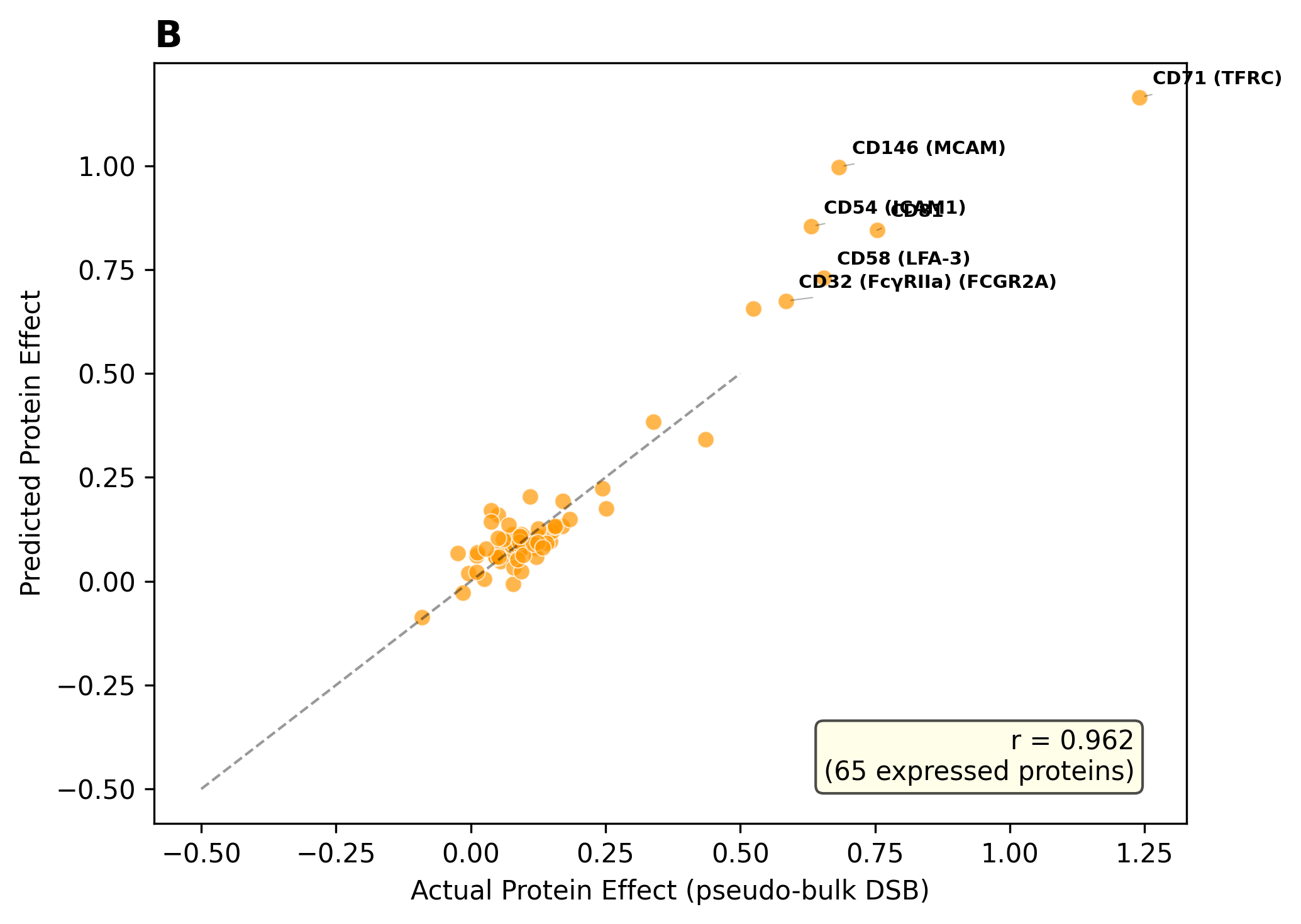}
\end{minipage}%
\hfill
\begin{minipage}[c]{0.56\textwidth}
\includegraphics[width=\textwidth]{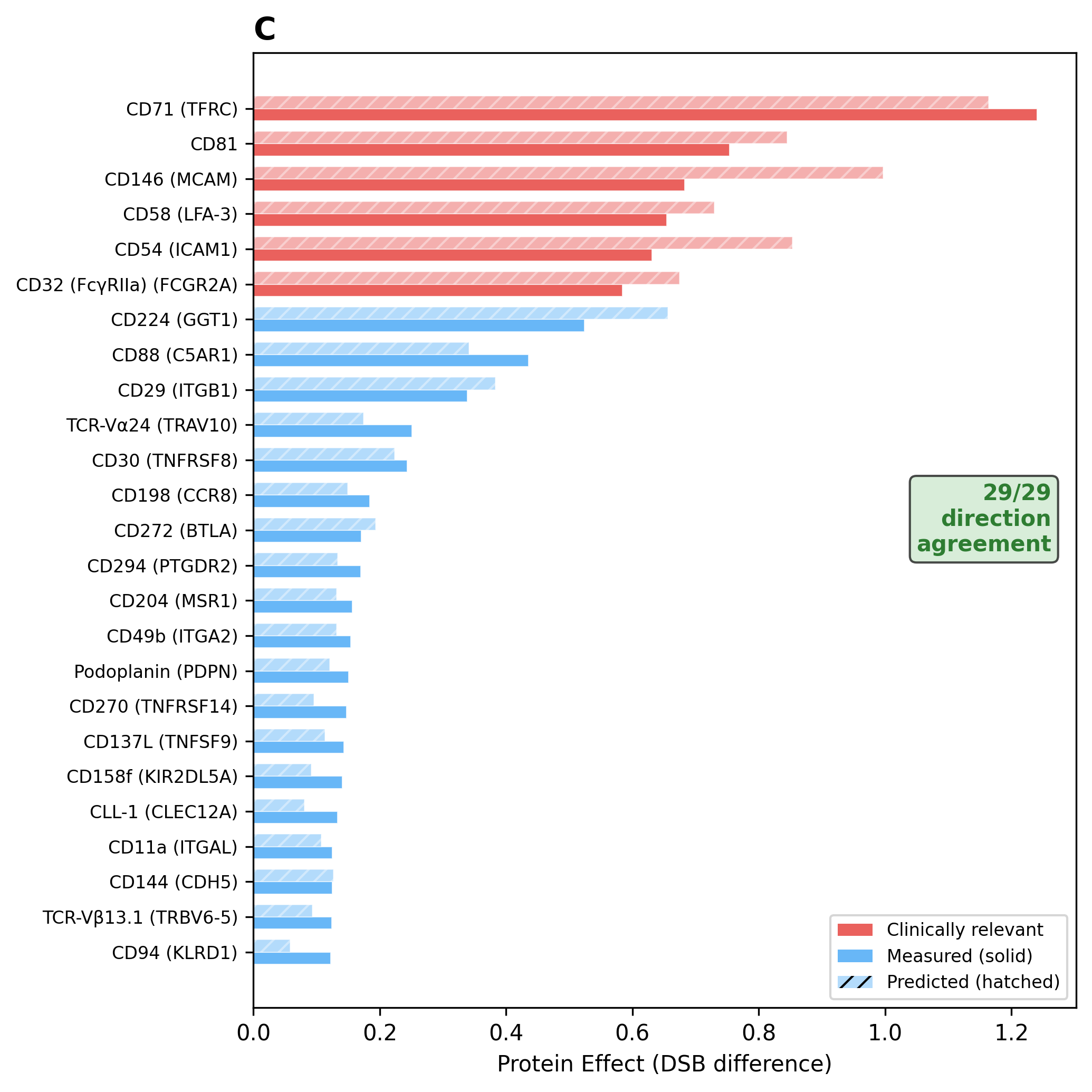}
\end{minipage}
\caption{\textbf{In silico Alemtuzumab side effect prediction.} \textbf{(A)} CD52 knockdown RNA prediction vs.\ actual (per-gene $r = 0.748$). \textbf{(B)} Protein prediction vs.\ actual for 65 expressed proteins ($r = 0.962$). \textbf{(C)} Side effect map: predicted vs.\ measured protein changes for 29 proteins with detectable effects (29/29 direction agreement, color-coded by clinical relevance).}
\label{fig:cd52}
\end{figure}

\textbf{Side effect map: 29/29 direction agreement.} Of 65 expressed proteins, 29 show detectable changes following CD52 knockdown (absolute DSB difference $> 0.1$). All 29 predictions match the observed direction of change (Figure~\ref{fig:cd52}C). The top changes (ranked by measured effect size) are: TFRC (CD71)$\uparrow$ (iron metabolism), CD81$\uparrow$ (B cell co-receptor), MCAM (CD146)$\uparrow$ (endothelial adhesion), CD58 (LFA-3)$\uparrow$ (immune reconstitution), ICAM1 (CD54)$\uparrow$ (inflammatory adhesion), and FCGR2A (CD32)$\uparrow$ (Fc receptor, autoimmunity link). Notably, while CD52 knockdown causes widespread mRNA decreases (Figure~\ref{fig:cd52}A), the experimentally measured surface proteins frequently change in the opposite direction: among the 27 expressed proteins with matched mRNA measurements in the STING-seq experimental data~\citep{morris2023}, the majority of those with observable mRNA changes ($|\text{log2FC}| > 0.01$) show opposite directions of change between measured mRNA and measured protein (8/12, 66.7\%; Appendix~\ref{app:rna_protein_divergence}), underscoring why protein-level prediction is essential for safety assessment.

\textbf{Systematic clinical validation.} We compare CDT-III's protein-level predictions against known Alemtuzumab side effects from clinical literature~\citep{ruck2022,thompson2010,devonshire2018} (Table~\ref{tab:side_effects}). Of 7 major known side effects, CDT-III recapitulates 5, providing mechanistic protein-level explanations for each. The 2 unmatched side effects (Goodpasture syndrome, listeria meningitis) involve pathogen-specific or tissue-specific mechanisms outside K562's biology.

\begin{table}[t]
\centering
\caption{\textbf{CDT-III predictions vs.\ known Alemtuzumab side effects.} Five of seven known side effects are recapitulated by CDT-III's protein predictions, without access to any clinical data.}
\label{tab:side_effects}
\small
\begin{tabular}{p{3.2cm}cp{4.2cm}c}
\toprule
Known side effect & Incidence & CDT-III predicted proteins & Match \\
\midrule
Autoimmune thyroid disease & 36--42\% & FCGR2A$\uparrow$, BAFF-R$\uparrow$, CD40$\uparrow$ & \ding{51} \\
ITP & 2--3\% & FCGR2A$\uparrow$ & \ding{51} \\
Infection risk & 66--77\% & CD58$\uparrow^{\star}$, TFRC$\uparrow^{\star}$ & \ding{51} \\
Infusion reaction & $>$90\% & ICAM1$\uparrow$, CD29$\uparrow$ & \ding{51} \\
Autoimmune nephropathy & 0.2--0.3\% & MCAM$\uparrow$ & \ding{51} \\
\midrule
Goodpasture syndrome & $<$0.1\% & --- & \ding{55} \\
Listeria meningitis & rare & --- & \ding{55} \\
\bottomrule
\end{tabular}
\begin{flushleft}
\footnotesize{$^{\star}$CD58$\uparrow$ and TFRC$\uparrow$ are compensatory responses to lymphocyte depletion, serving as molecular markers of the immunocompromised state rather than direct causes of infection. Note that gradient analysis yields TFRC$\downarrow$, reflecting immediate perturbation sensitivity rather than the compensatory response captured by direct prediction.}
\end{flushleft}
\end{table}

\textbf{Gradient analysis without perturbation data.} The gradient analysis results above were obtained without any CD52 perturbation data---only the non-targeting control (NTC) mean expression and CD52's Enformer DNA embedding were provided as input, and the protein-level side effect profile was derived entirely from the trained model's learned weights via backpropagation ($\partial \hat{y}_{\text{Prot},j} / \partial x_{\text{RNA},g}$). To validate this approach, we compared gradients computed from NTC mean input against those computed from actual CD52-perturbed cells: on 65 expressed proteins, the two approaches show Pearson $r = 0.939$ and $96.9\%$ direction agreement (Appendix~\ref{app:ntc_gradient}). This confirms that gradient analysis can screen novel drug targets without perturbation data: any gene within the model's 2,361-gene vocabulary can be evaluated for potential side effects without conducting a new perturbation experiment.

\textbf{Novel predictions.} CDT-III also generates testable hypotheses not yet reported in the Alemtuzumab literature. Gradient analysis predicts B2M$\downarrow$, which would reduce MHC class I surface expression and potentially enable immune evasion. Direct prediction identifies CD81$\uparrow$ during B cell rebound, suggesting a novel biomarker for monitoring autoimmune risk post-treatment.

\vspace{-2mm}
\section{Discussion}
\vspace{-2mm}
\textbf{Multi-task regularization as a general principle.}
Adding protein prediction improves RNA prediction ($r = 0.804 \to 0.843$)---a non-trivial result, as multi-task learning often degrades individual task performance~\citep{crawshaw2020}. Protein supervision provides complementary gradient signals that sharpen shared representations, a principle applicable to any hierarchical multi-task system.

\textbf{Why two stages, not one.}
The single-stage VCE failure (Table~\ref{tab:ablation}) illustrates a tension between flexibility and transfer in multi-modal architectures. Adding modalities to a joint fusion layer breaks weight compatibility. The two-stage design resolves this by decomposing fusion along the causal structure of the data---a principle that may generalize to other multi-modal systems where modalities follow a natural hierarchy.

\textbf{Downstream supervision sharpens upstream interpretability.}
The $+30\%$ CTCF enrichment improvement is unexpected: protein data never enters the DNA pathway directly, yet protein loss gradients sharpen DNA-level attention toward genomic positions relevant to protein-level phenotypes. This sharpening operates at single-locus resolution---the model selectively attends to CTCF sites with confirmed 3D promoter contacts while ignoring nearby sites that lack such contacts (Appendix~\ref{app:genome_tracks})---suggesting a broader principle: in hierarchical systems, supervision at the most distal output can improve the fidelity of learned internal structure at every upstream layer.

\textbf{In silico pharmacology and the necessity of protein-level prediction.}
Gradient analysis produces concordant results from perturbed and unperturbed cells ($r = 0.939$; Appendix~\ref{app:ntc_gradient}), enabling scalable in silico drug screening of arbitrary targets. Critically, analysis of the experimentally measured mRNA and protein responses in the STING-seq data~\citep{morris2023} reveals that the majority of genes with observable mRNA changes show opposite protein-level changes (66.7\% at $|\text{log2FC}| > 0.01$, rising to 87.5\% at $|\text{log2FC}| > 0.02$; Appendix~\ref{app:rna_protein_divergence}). This exposes a fundamental limitation of RNA-only models~\citep{roohani2024gears,lotfollahi2023cpa}: the implicit assumption that protein follows mRNA direction fails for the majority of genes with measurable mRNA responses. CDT-III's VCE-C cross-attention learns to decouple these responses, achieving protein $r = 0.962$ despite pervasive direction discordance---a capability architecturally impossible in single-modality models.

\textbf{Limitations.}
All experiments use K562 cells, reflecting the scarcity of multi-modal perturbation datasets (Section~\ref{sec:setup}); the architecture applies without modification to other cell types as such data become available.
CRISPRi knockdown approximates but does not recapitulate antibody drug action (CRISPRi suppresses transcription; Alemtuzumab depletes via ADCC/CDC).
Per-protein cell-level $r = 0.28$ limits single-cell applications. Gradient-based interpretability (correlation between gradient-derived regulatory importance and CTCF binding) achieves $r = 0.443$ (vs.\ CDT-II's $r = 0.83$), though only 2 held-out genes with sufficient CTCF sites are testable. Hi-C permutation $P = 0.088$ is marginal, though parametric tests are significant ($P = 0.020$).

\textbf{Broader impact.}
In silico drug safety screening could prioritize candidates and identify adverse effects early. More broadly, the mRNA--protein direction divergence reported here suggests that CRISPRi screens relying solely on RNA readouts may miss or mischaracterize protein-level phenotypes critical for drug development.

\vspace{-2mm}
\section{Conclusion}
\vspace{-2mm}
The interpretability gap in biological AI can be closed by aligning model architecture with the cell's own logic. CDT-III connects all three layers of the central dogma (DNA $\to$ RNA $\to$ Protein) in a single interpretable architecture---to our knowledge, the first to do so for perturbation response prediction---showing that prediction and mechanistic understanding need not be separate goals. Each attention map corresponds to a specific biological process, and every prediction can be traced through a mechanistic pathway.

Three principles extend beyond biology: downstream supervision improves upstream representation, multi-modal fusion should respect causal data structure, and mechanism-oriented AI can bridge basic science to clinical application. The path to trustworthy AI in biomedicine lies in architectures that reveal the mechanisms underlying their predictions.

\begin{ack}
The author thanks Morris et al.\ for making the STING-seq v2 dataset publicly available, the developers of Enformer, and the open-source communities behind PyTorch and related tools. Claude (Anthropic) was used for manuscript editing. This work received no external funding.
\end{ack}

\bibliographystyle{plainnat}

\appendix

\section{CDT-II Architecture}
\label{app:cdtii}

\begin{figure}[h]
\centering
\includegraphics[width=0.7\textwidth]{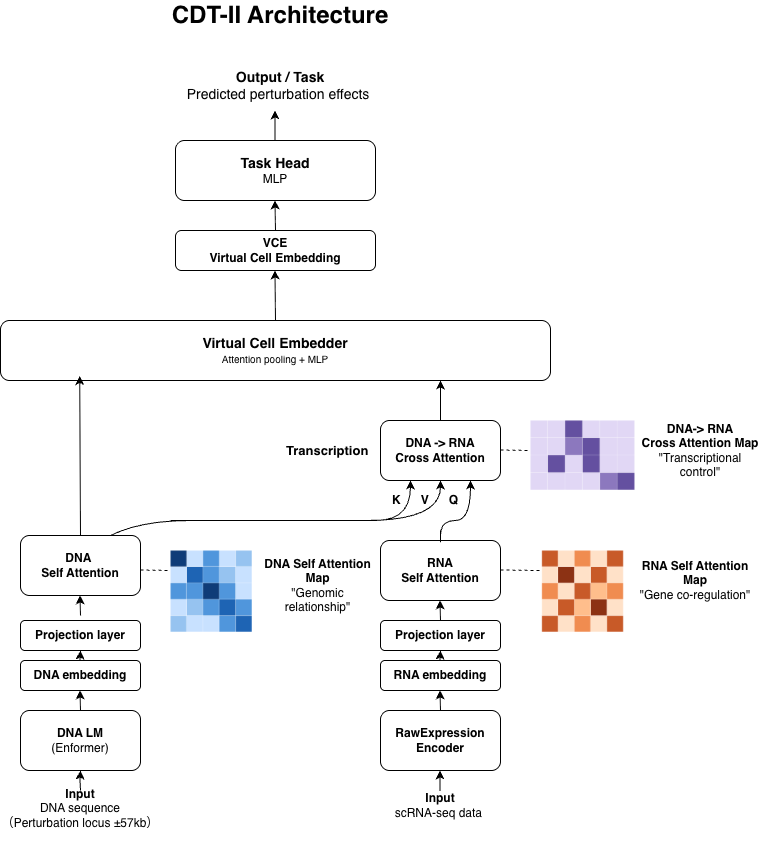}
\caption{\textbf{CDT-II architecture.} The model mirrors the central dogma: DNA self-attention captures genomic relationships within a $\pm$57 kb window, RNA self-attention captures gene co-regulation, and cross-attention models transcriptional control. A Virtual Cell Embedder integrates both modalities to predict perturbation effects. CDT-III's VCE-N preserves this architecture exactly, enabling 100\% weight transfer.}
\label{fig:cdtii_arch}
\end{figure}

CDT-III's VCE-N is architecturally identical to CDT-II. Figure~\ref{fig:cdtii_arch} shows the CDT-II architecture from which all VCE-N weights are transferred.

\section{Detailed Architecture}
\label{app:architecture}

\subsection{VCE-N (Nuclear --- Transcription Stage)}

VCE-N is identical to CDT-II~\citep{ota2026cdtii}:

\begin{itemize}
    \item \textbf{DNA pathway}: Linear projection ($3072 \to 512$) + LayerNorm $\to$ 2 self-attention layers ($h=8$, FFN $= 2048$, GELU, dropout 0.3)
    \item \textbf{RNA pathway}: RawExpressionEncoder (gene identity embeddings $d=512$ + expression value projection) $\to$ 1 self-attention layer
    \item \textbf{Cross-attention}: RNA queries, DNA keys/values ($h=8$)
    \item \textbf{VCE pooling}: 4-head attention pooling per modality $\to$ concatenate $\to$ MLP ($d \times 2 \to d \times 2 \to d$) $\to$ $\mathbf{e}_{\text{RNA}} \in \mathbb{R}^{512}$
    \item \textbf{RNA task head}: 2-layer MLP ($512 \to 1024 \to 2361$)
\end{itemize}

\subsection{VCE-C (Cytosolic --- Translation Stage)}

VCE-C is new in CDT-III:

\begin{itemize}
    \item \textbf{Protein pathway}: RawExpressionEncoder ($n=189$, $d=512$) $\to$ 1 self-attention layer
    \item \textbf{RNA$\to$Protein cross-attention}: Protein queries, RNA keys/values ($h=8$)
    \item \textbf{VCE pooling}: Takes $\mathbf{e}_{\text{RNA}}$ from VCE-N + protein representation $\to$ MLP ($d \times 2 \to d \times 2 \to d$) $\to$ $\mathbf{e}_{\text{Prot}} \in \mathbb{R}^{512}$
    \item \textbf{Protein task head}: 2-layer MLP ($512 \to 1024 \to 189$)
\end{itemize}

\section{Ablation: Single-Stage vs.\ Two-Stage VCE}
\label{app:ablation}

We trained five single-stage VCE variants that fuse DNA, RNA, and protein embeddings jointly, varying the protein loss weight $\lambda$ (Table~\ref{tab:ablation}).

\begin{table}[h]
\centering
\caption{\textbf{Single-stage vs.\ two-stage VCE.} Per-cell correlation ($r$) on the validation set. All models transfer CDT-II encoder and attention weights; the critical difference is the VCE fusion layer. Single-stage architectures modify fusion dimensions ($d \times 2 \to d \times 3$), preventing fusion weight transfer and causing RNA prediction to collapse despite encoder transfer. The two-stage architecture preserves full weight compatibility including fusion. $^*$Protein $r$ values for single-stage models are inflated by 124/189 non-expressed proteins predicted as $\sim$0.}
\label{tab:ablation}
\begin{tabular}{llcccc}
\toprule
& Architecture & $\lambda$ & RNA $r$ & Protein $r$ & Fusion transfer \\
\midrule
CDT-II & RNA only & --- & 0.64 & --- & --- \\
\midrule
Exp A & 1-Stage & 1.0 & 0.19 & 0.76$^*$ & \ding{55} \\
Exp B & 1-Stage & 0.11 & 0.24 & 0.77$^*$ & \ding{55} \\
Exp C & 1-Stage & 0.05 & 0.35 & 0.81$^*$ & \ding{55} \\
Exp D & 1-Stage & 0.01 & 0.37 & 0.80$^*$ & \ding{55} \\
Exp E & 1-Stage ($\lambda$ warmup) & 0.001$\to$0.01 & 0.30 & 0.21 & \ding{55} \\
\midrule
\textbf{CDT-III} & \textbf{2-Stage VCE} & \textbf{0.1} & \textbf{0.64} & \textbf{0.43} & \textbf{\ding{51}} \\
\bottomrule
\end{tabular}
\end{table}

All single-stage models transfer CDT-II's encoder and attention weights (DNA projector, self-attention layers, RNA encoder, cross-attention), yet fail to reach CDT-II's RNA performance ($r = 0.64$), with the best achieving only $r = 0.37$. The failure traces to a single architectural component: modifying VCE fusion dimensions from $d \times 2$ to $d \times 3$ prevents loading CDT-II's pre-trained fusion weights, forcing this critical layer to learn from random initialization. Even with curriculum learning ($\lambda$ warmup, Exp~E), RNA performance does not recover---demonstrating that the VCE fusion layer is the bottleneck, not the encoders.

The two-stage VCE resolves this by preserving VCE-N's dimensions exactly, enabling 100\% weight transfer including the fusion layer. VCE-C is added as a new module with its own fusion layer ($d \times 2 \to d \times 2 \to d$), initialized from scratch but trained without disrupting VCE-N. The per-cell protein $r = 0.43$ (on 65 expressed proteins only) may appear modest, but pseudo-bulk (per-gene mean) evaluation yields $r = 0.969$, revealing that CDT-III accurately captures perturbation-level protein response patterns despite single-cell noise.

\section{Gradient Flow}
\label{app:gradient}

The two-stage design creates two distinct gradient pathways:

\begin{align}
\frac{\partial \mathcal{L}_{\text{RNA}}}{\partial \theta_{\text{VCE-N}}} &: \text{Direct (same as CDT-II)} \\
\frac{\partial \mathcal{L}_{\text{Prot}}}{\partial \theta_{\text{VCE-N}}} &: \text{Through } \mathbf{e}_{\text{RNA}} \to \text{VCE-C} \to \mathcal{L}_{\text{Prot}} \text{ (new in CDT-III)}
\end{align}

The protein loss gradient provides additional supervision to VCE-N, explaining why CTCF enrichment improves from $6.6\times$ to $8.59\times$.

\section{Attention Map Inventory}
\label{app:attention}

CDT-III produces six attention maps, compared to CDT-II's three:

\begin{table}[h]
\centering
\caption{\textbf{Attention map inventory.} CDT-III produces six interpretable attention maps, each corresponding to a specific biological process. Four are inherited from CDT-II; two are new.}
\label{tab:attention_inventory}
\begin{tabular}{llll}
\toprule
Attention Map & Shape & Biological Meaning & Source \\
\midrule
DNA Self-Attention (L0) & $[8, 896, 896]$ & Genomic interactions & CDT-II \\
DNA Self-Attention (L1) & $[8, 896, 896]$ & Higher-order genomic & CDT-II \\
RNA Self-Attention & $[8, 2361, 2361]$ & Gene co-regulation & CDT-II \\
DNA$\to$RNA Cross-Attention & $[8, 2361, 896]$ & Transcriptional control & CDT-II \\
Protein Self-Attention & $[8, 189, 189]$ & Protein co-regulation & \textbf{CDT-III} \\
RNA$\to$Protein Cross-Attention & $[8, 189, 2361]$ & Translational control & \textbf{CDT-III} \\
\bottomrule
\end{tabular}
\end{table}

\section{Per-Gene CTCF Enrichment}
\label{app:ctcf}

\begin{table}[H]
\centering
\small
\caption{CTCF enrichment for all 28 genes with ENCODE K562 ChIP-seq data. 14/27 genes with CTCF sites show perfect 10.0$\times$ enrichment (all CTCF sites in top 10\% attention bins).}
\begin{tabular}{lccclccc}
\toprule
Gene & CTCF & Top 10\% & Enrich. & Gene & CTCF & Top 10\% & Enrich. \\
\midrule
CD82 & 7 & 7 & 10.0$\times$ & SLC3A2 & 6 & 5 & 8.33$\times$ \\
CD69 & 7 & 7 & 10.0$\times$ & TNFSF9 & 5 & 4 & 8.00$\times$ \\
CD40 & 9 & 9 & 10.0$\times$ & CD44 & 4 & 3 & 7.50$\times$ \\
ITGB1 & 2 & 2 & 10.0$\times$ & CD58 & 4 & 3 & 7.50$\times$ \\
ENTPD1 & 2 & 2 & 10.0$\times$ & CD55 & 4 & 3 & 7.50$\times$ \\
KLRG1 & 3 & 3 & 10.0$\times$ & LGALS3 & 4 & 3 & 7.50$\times$ \\
KLRD1 & 2 & 2 & 10.0$\times$ & ITGAE & 4 & 3 & 7.50$\times$ \\
TNFSF13B & 3 & 3 & 10.0$\times$ & FCGR2A & 2 & 1 & 5.00$\times$ \\
B2M & 4 & 4 & 10.0$\times$ & CD83 & 5 & 2 & 4.00$\times$ \\
CD19 & 4 & 4 & 10.0$\times$ & IL21R & 6 & 2 & 3.33$\times$ \\
NECTIN2 & 4 & 4 & 10.0$\times$ & MSR1 & 0 & 0 & N/A \\
EPCAM & 4 & 4 & 10.0$\times$ & & & & \\
TFRC & 2 & 2 & 10.0$\times$ & & & & \\
GFI1B & 3 & 3 & 10.0$\times$ & & & & \\
\midrule
ICAM1 & 8 & 7 & 8.75$\times$ & \textbf{Mean} & & & \textbf{8.59$\times$} \\
CD52 & 7 & 6 & 8.57$\times$ & & & & \\
CD81 & 7 & 6 & 8.57$\times$ & & & & \\
\bottomrule
\end{tabular}
\label{tab:ctcf_full}
\end{table}

\section{1-Stage VCE Ablation Details}
\label{app:ablation_full}

\begin{table}[H]
\centering
\caption{Full results for all five single-stage VCE experiments.}
\begin{tabular}{llccccl}
\toprule
Exp & $\lambda$ & Strategy & RNA $r$ & Prot $r$ & Epochs & Notes \\
\midrule
A & 1.0 & Fixed & 0.19 & 0.76 & 148 & Protein-dominated loss \\
B & 0.11 & Fixed & 0.24 & 0.77 & 209 & Equal-weight by variance \\
C & 0.05 & Fixed & 0.35 & 0.81 & 386 & Best single-stage protein \\
D & 0.01 & Fixed & 0.37 & 0.80 & 472 & Best single-stage RNA \\
E & 0.001$\to$0.01 & Warmup & 0.30 & 0.21 & $\sim$300 & Curriculum failed \\
\bottomrule
\end{tabular}
\label{tab:ablation_full}
\end{table}

\textbf{Root cause analysis.} All single-stage experiments share the same fundamental limitation: the VCE fusion layer requires $d \times 3$ input dimensions (DNA + RNA + Protein pooled representations), compared to CDT-II's $d \times 2$. This dimensional mismatch prevents loading CDT-II's pre-trained fusion weights, forcing the fusion layer to learn from random initialization. Since CDT-II required $>$500 epochs to reach $r = 0.57$ from scratch, the single-stage fusion layer cannot recover to comparable quality within feasible training time. The two-stage VCE avoids this entirely by keeping VCE-N dimensions unchanged.

\section{CD52 Side Effect Prediction Details}
\label{app:cd52}

\begin{table}[H]
\centering
\small
\caption{Top 20 proteins predicted to change upon CD52 knockdown (gradient-based analysis).}
\begin{tabular}{clcl}
\toprule
Rank & Protein (Gene) & Gradient & Category \\
\midrule
1 & CD32 (FCGR2A) & $+0.059\uparrow$ & Fc receptor / ADCC \\
2 & CD146 (MCAM) & $+0.044\uparrow$ & Endothelial adhesion \\
3 & CD54 (ICAM1) & $+0.043\uparrow$ & Inflammatory adhesion \\
4 & CD81 (CD81) & $+0.031\uparrow$ & B cell co-receptor \\
5 & CD58 (CD58) & $+0.030\uparrow$ & T cell activation (LFA-3) \\
6 & CD224 (GGT1) & $+0.030\uparrow$ & Oxidative stress \\
7 & CD29 (ITGB1) & $+0.019\uparrow$ & Cell adhesion \\
8 & Ig$\kappa$ (IGKC) & $+0.010\uparrow$ & B cell / antibody \\
9 & TCR V$\gamma$9 (TRGV9) & $+0.010\uparrow$ & $\gamma\delta$ T cell \\
10 & CD71 (TFRC) & $-0.009\downarrow$ & Iron metabolism \\
11 & Ig$\lambda$ (IGLC1) & $+0.008\uparrow$ & B cell / antibody \\
12 & CD11c (ITGAX) & $+0.008\uparrow$ & Dendritic cell \\
13 & CD40 (CD40) & $+0.007\uparrow$ & B cell activation \\
14 & TSLPR (CRLF2) & $+0.007\uparrow$ & Autoimmunity \\
15 & CD5 (CD5) & $+0.006\uparrow$ & T/B1 cell \\
16 & CD144 (CDH5) & $+0.006\uparrow$ & Endothelial barrier \\
17 & $\beta_2$M (B2M) & $-0.006\downarrow$ & MHC class I \\
18 & CD268 (TNFRSF13C) & $+0.006\uparrow$ & BAFF-R / B cell survival \\
19 & CD198 (CCR8) & $+0.006\uparrow$ & Treg migration \\
20 & CD294 (PTGDR2) & $+0.005\uparrow$ & Th2 / allergy \\
\bottomrule
\end{tabular}
\label{tab:cd52_full}
\end{table}

\section{Hi-C Validation Details}
\label{app:hic}

\begin{table}[H]
\centering
\small
\caption{Per-gene Hi-C contact ratios for all 25 genes with sufficient coverage (4DN K562 data, 5 kb resolution). Three genes excluded due to zero Hi-C records: \textit{B2M}, \textit{TFRC}, \textit{GFI1B}. Mean ratio $= 1.30\times$; 16/25 genes show ratio $> 1$ ($P = 0.020$, Wilcoxon signed-rank).}
\begin{tabular}{lcccc}
\toprule
Gene & Top 20 Hi-C & Random 20 Hi-C & Ratio & Note \\
\midrule
CD55 & 156.4 & 56.8 & $2.75\times$ & Complement regulator \\
CD82 & 66.2 & 29.0 & $2.28\times$ & Tetraspanin \\
IL21R & 68.6 & 31.2 & $2.20\times$ & IL-21 receptor \\
FCGR2A & 93.8 & 45.1 & $2.08\times$ & Fc receptor \\
CD83 & 71.5 & 36.0 & $1.99\times$ & Dendritic cell \\
ITGB1 & 51.3 & 29.9 & $1.72\times$ & Integrin $\beta$1 \\
ITGAE & 74.7 & 49.0 & $1.53\times$ & Integrin $\alpha$E \\
CD44 & 42.1 & 28.3 & $1.49\times$ & Cell adhesion \\
SLC3A2 & 47.9 & 34.4 & $1.39\times$ & Amino acid transporter \\
MSR1 & 48.6 & 35.4 & $1.37\times$ & Scavenger receptor \\
EPCAM & 30.7 & 23.2 & $1.32\times$ & Epithelial adhesion \\
TNFSF13B & 18.0 & 13.8 & $1.30\times$ & BAFF \\
NECTIN2 & 75.4 & 57.9 & $1.30\times$ & Cell--cell adhesion \\
LGALS3 & 30.3 & 24.4 & $1.24\times$ & Galectin-3 \\
KLRD1 & 31.8 & 28.5 & $1.12\times$ & NK receptor \\
CD40 & 30.4 & 28.1 & $1.08\times$ & B cell activation \\
\midrule
CD69 & 36.5 & 37.4 & $0.97\times$ & Early activation \\
ICAM1 & 56.6 & 60.0 & $0.94\times$ & Inflammatory adhesion \\
KLRG1 & 30.3 & 38.3 & $0.79\times$ & NK/T inhibitory \\
ENTPD1 & 21.4 & 29.7 & $0.72\times$ & ATP metabolism \\
CD58 & 27.4 & 39.4 & $0.70\times$ & LFA-3 \\
CD52 & 35.7 & 53.5 & $0.67\times$ & CAMPATH-1 antigen \\
CD81 & 26.1 & 41.9 & $0.62\times$ & Tetraspanin \\
TNFSF9 & 27.0 & 48.7 & $0.56\times$ & 4-1BB ligand \\
CD19 & 20.8 & 61.4 & $0.34\times$ & B cell marker \\
\midrule
\textbf{Mean} & & & $\mathbf{1.30\times}$ & \textbf{16/25 $> 1$} \\
\bottomrule
\end{tabular}
\label{tab:hic_full}
\end{table}

\section{Gradient Analysis: NTC Mean Validation}
\label{app:ntc_gradient}

To validate that gradient analysis can predict side effects without perturbation data, we compared two gradient computation approaches for CD52: (1)~using actual CD52-perturbed cells as input (as in Section~\ref{sec:cd52}), and (2)~using only the non-targeting control (NTC) mean expression and NTC mean protein levels---i.e., no CD52 perturbation data whatsoever. Table~\ref{tab:ntc_validation} shows that the two approaches produce highly concordant results, confirming that gradient analysis can screen novel drug targets from unperturbed baseline data alone.

\begin{table}[H]
\centering
\caption{\textbf{Gradient analysis validation: NTC mean vs.\ perturbation cell input.} Concordance between protein-level gradients ($\partial \hat{y}_{\text{Prot},j} / \partial x_{\text{RNA, CD52}}$) computed using NTC mean input (no perturbation data) vs.\ actual CD52-perturbed cell input.}
\label{tab:ntc_validation}
\begin{tabular}{lcc}
\toprule
Metric & All proteins (189) & Expressed proteins (65) \\
\midrule
Pearson $r$ & 0.913 & 0.939 \\
Spearman $\rho$ & 0.752 & 0.865 \\
Direction agreement & 145/189 (76.7\%) & 63/65 (96.9\%) \\
Top 20 direction agreement & 20/20 (100\%) & --- \\
\bottomrule
\end{tabular}
\end{table}

\section{mRNA--Protein Response Divergence}
\label{app:rna_protein_divergence}

To comprehensively assess the relationship between mRNA and surface protein responses to CD52 knockdown, we analyzed the experimentally measured values from the Morris et al.\ STING-seq CRISPRi dataset~\citep{morris2023}---not CDT-III predictions. We computed mRNA log2 fold changes from the full transcriptome (36,601 genes; CD52 knockdown vs.\ non-targeting control, 1,369 vs.\ 39,101 cells) and used DSB-normalized protein effects (131 CD52 cells, consistent with all other protein analyses in this paper). Matching the 65 detectably expressed proteins to their corresponding gene symbols yielded 27 unique gene--protein pairs (26 with nonzero changes).

Table~\ref{tab:divergence_sensitivity} shows a sensitivity analysis across mRNA change thresholds. Among all 26 pairs with nonzero changes, 16 (61.5\%) show opposite directions---mRNA decreases while protein increases. As the threshold increases to focus on genes with more reliable mRNA changes, the divergence rate rises further: at $|\text{log2FC}| > 0.01$, 8/12 (66.7\%) show opposite directions; at $|\text{log2FC}| > 0.02$, 7/8 (87.5\%); and at $|\text{log2FC}| > 0.05$, 4/4 (100\%). This pattern demonstrates that for genes where mRNA responds measurably to perturbation, the protein-level response predominantly moves in the opposite direction.

\begin{table}[H]
\centering
\caption{\textbf{Sensitivity analysis of mRNA--protein direction divergence.} For each mRNA change threshold, we report the number of gene--protein pairs with opposite vs.\ same direction of change following CD52 knockdown. mRNA: full transcriptome log2FC; protein: DSB-normalized effects (65 expressed proteins, 27 matched pairs).}
\label{tab:divergence_sensitivity}
\small
\begin{tabular}{lcccc}
\toprule
$|\text{log2FC}|$ threshold & $N$ genes & Opposite & Same & \% Opposite \\
\midrule
$> 0$ (all nonzero) & 26 & 16 & 10 & 61.5\% \\
$> 0.005$ & 16 & 10 & 6 & 62.5\% \\
$> 0.01$ & 12 & 8 & 4 & 66.7\% \\
$> 0.02$ & 8 & 7 & 1 & 87.5\% \\
$> 0.05$ & 4 & 4 & 0 & 100.0\% \\
\bottomrule
\end{tabular}
\end{table}

\begin{figure}[H]
\centering
\includegraphics[width=\textwidth]{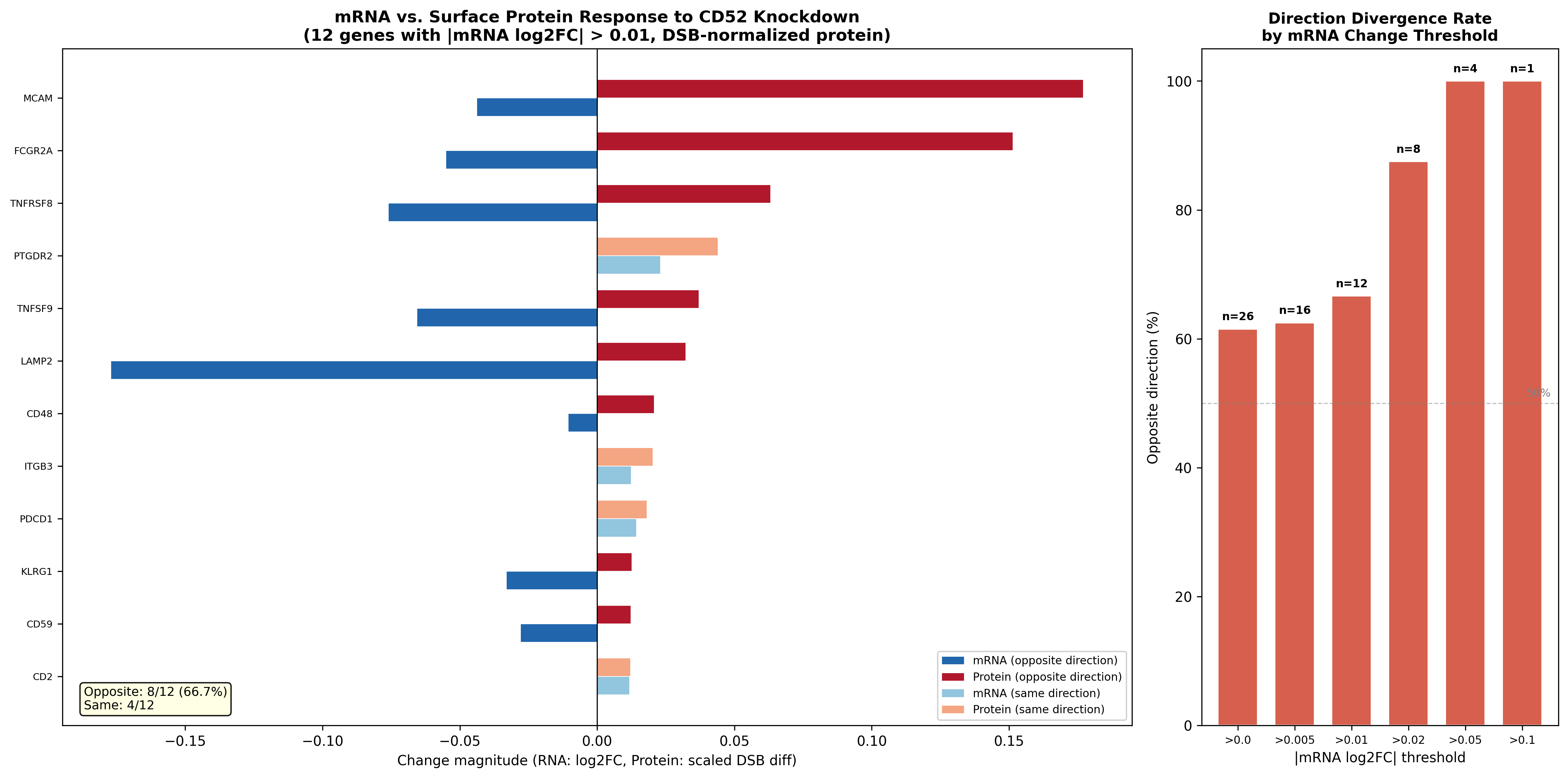}
\caption{\textbf{mRNA vs.\ surface protein response divergence after CD52 knockdown.} \textbf{Left:} Paired bar chart for 12 genes with $|\text{mRNA log2FC}| > 0.01$ (DSB-normalized protein), showing mRNA change (blue) and protein change (red, scaled) for each gene. Dark colors indicate opposite-direction pairs (8/12, 66.7\%); light colors indicate same-direction pairs. The dominant pattern---mRNA decreases while surface protein increases---is consistent with post-translational regulation. \textbf{Right:} Sensitivity analysis showing the percentage of gene--protein pairs with opposite directions at increasing mRNA change thresholds. The divergence rate rises from 61.5\% to 100\% as the threshold increases, indicating that genes with the strongest mRNA responses show the most pronounced protein-level discordance.}
\label{fig:divergence}
\end{figure}

Figure~\ref{fig:divergence} shows the per-gene comparison for the 12 genes with $|\text{log2FC}| > 0.01$. Dark-colored bars indicate gene--protein pairs with opposite directions of change; light-colored bars indicate same-direction pairs. The dominant pattern is mRNA decrease with protein increase: for example, ICAM1 mRNA decreases ($-0.124$ log2FC) while its surface protein increases, consistent with post-translational compensation. These discordances cannot be captured by mRNA-level analysis alone.

Despite this divergence, CDT-III correctly predicts both the mRNA decrease and the protein increase for these genes. This implies that the model has learned to decouple mRNA-level and protein-level responses---likely through the RNA$\to$Protein cross-attention in VCE-C, which allows each protein to selectively attend to gene expression patterns that drive its surface-level change, independent of the target gene's own mRNA direction. Analyzing which genes each protein attends to in this cross-attention layer could reveal the compensatory regulatory mechanisms underlying the observed mRNA--protein divergence, a direction we leave for future work.

\section{Cross-Attention Genome Tracks}
\label{app:genome_tracks}

\begin{figure}[H]
\centering
\includegraphics[width=0.9\textwidth]{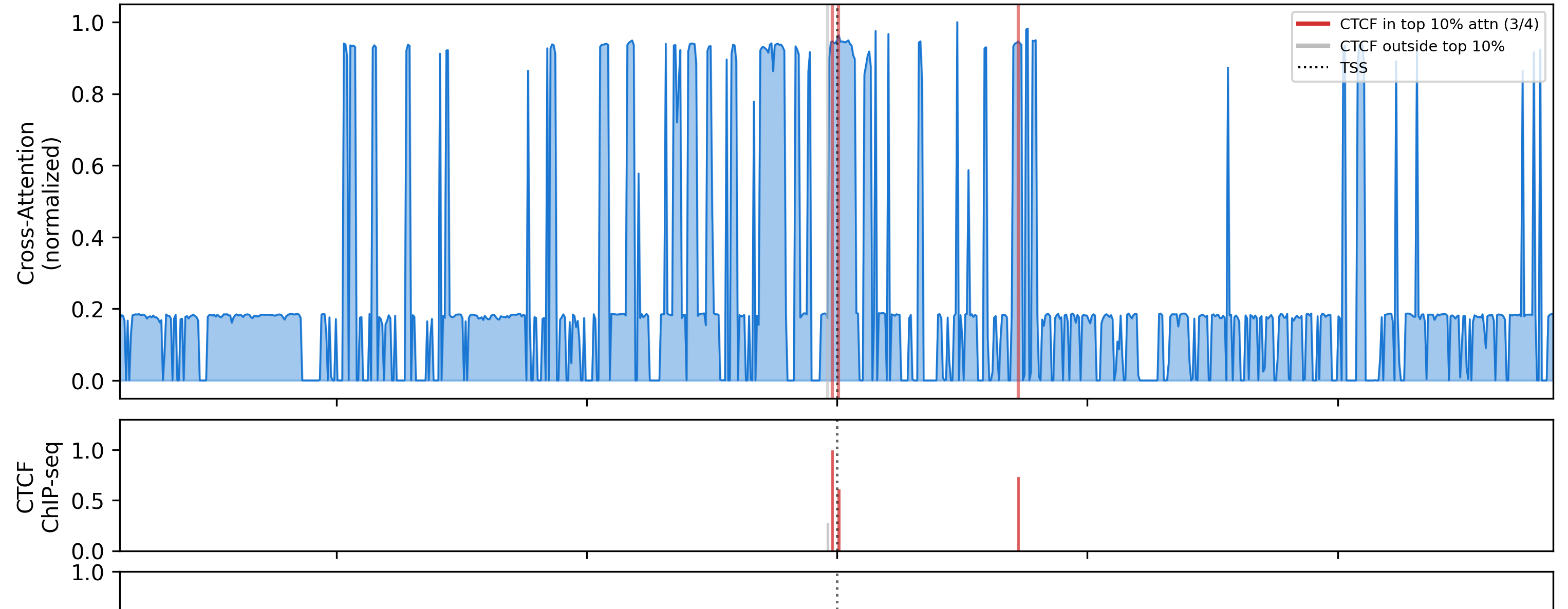}
\caption{\textbf{CD55 cross-attention genome track.} DNA$\to$RNA cross-attention profile (top) aligned with CTCF ChIP-seq peaks (bottom) across the $\pm$57 kb Enformer window. Red lines: CTCF sites in top 10\% attention (3/4). TSS indicated by dotted line. Hi-C contact ratio: $2.75\times$ (Table~\ref{tab:hic_full}).}
\label{fig:genome_cd55}
\end{figure}

\begin{figure}[H]
\centering
\includegraphics[width=0.9\textwidth]{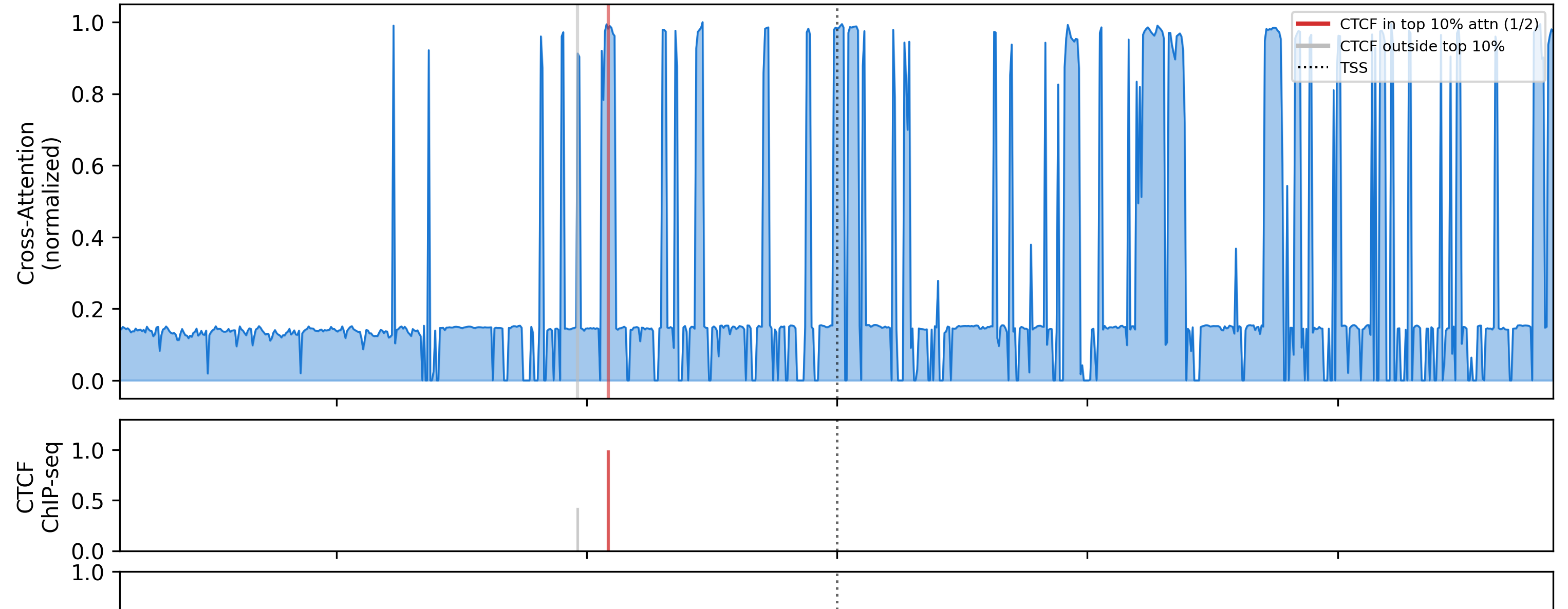}
\caption{\textbf{FCGR2A cross-attention genome track.} Same format as Figure~\ref{fig:genome_cd55}. CTCF site in top 10\% attention (1/2) located upstream of TSS, consistent with distal regulatory contact. Hi-C contact ratio: $2.08\times$.}
\label{fig:genome_fcgr2a}
\end{figure}

To illustrate what CTCF enrichment means at the individual gene level, Figure~\ref{fig:genome_cd55} and Figure~\ref{fig:genome_fcgr2a} show the DNA$\to$RNA cross-attention profile aligned with CTCF ChIP-seq peaks for two representative genes. Red vertical lines indicate CTCF binding sites that fall within the top 10\% of attention bins; gray lines indicate CTCF sites outside the top 10\%.

For CD55 (Hi-C contact ratio $2.75\times$), 3 of 4 CTCF sites fall within the highest-attention regions, demonstrating that the model has learned to attend to positions of known insulator function. For FCGR2A ($2.08\times$), the single high-attention CTCF site lies upstream of the TSS, consistent with a distal regulatory element mediating 3D chromatin contact.

These genome-level views confirm that the aggregate CTCF enrichment statistic ($8.59\times$; Table~\ref{tab:ctcf}) reflects genuine, position-specific alignment between learned attention and known regulatory architecture. Notably, the model does not attend to all CTCF sites equally: it selectively focuses on those that mediate 3D contacts with the promoter, distinguishing functionally relevant insulators from those not involved in regulating a given gene---a discrimination that emerges entirely from perturbation prediction training, without any chromatin structure supervision.

\newpage
\section*{NeurIPS Paper Checklist}

\begin{enumerate}

\item {\bf Claims}
    \item[] Question: Do the main claims made in the abstract and introduction accurately reflect the paper's contributions and scope?
    \item[] Answer: \answerYes{}
    \item[] Justification: All claims (multi-task regularization, interpretability improvement, in silico pharmacology) are supported by experimental results in Sections~5.2--5.4.

\item {\bf Limitations}
    \item[] Question: Does the paper discuss the limitations of the work performed by the authors?
    \item[] Answer: \answerYes{}
    \item[] Justification: Section~6 (Limitations) discusses K562-only evaluation, cell-level protein $r = 0.28$, gradient interpretability $r = 0.443$ on 2 genes, CRISPRi vs.\ antibody action, and marginal Hi-C permutation $P$.

\item {\bf Theory assumptions and proofs}
    \item[] Question: For each theoretical result, does the paper provide the full set of assumptions and a complete (and correct) proof?
    \item[] Answer: \answerNA{}
    \item[] Justification: This is an empirical paper with no theoretical results.

\item {\bf Experimental result reproducibility}
    \item[] Question: Does the paper fully disclose all the information needed to reproduce the main experimental results of the paper to the extent that it affects the main claims and/or conclusions of the paper (regardless of whether the code and data are provided or not)?
    \item[] Answer: \answerYes{}
    \item[] Justification: Full architecture details (Section~3, Appendix~B), hyperparameters (Sections~3.3--3.4), data splits (Section~4), and training procedure are provided. Code will be released upon publication.

\item {\bf Open access to data and code}
    \item[] Question: Does the paper provide open access to the data and code, with sufficient instructions to faithfully reproduce the main experimental results, as described in supplemental material?
    \item[] Answer: \answerYes{}
    \item[] Justification: The dataset is publicly available (GEO: GSE171452). Code will be released upon publication.

\item {\bf Experimental setting/details}
    \item[] Question: Does the paper specify all the training and test details (e.g., data splits, hyperparameters, how they were chosen, type of optimizer) necessary to understand the results?
    \item[] Answer: \answerYes{}
    \item[] Justification: Data splits (Section~4), two-phase training strategy with learning rates and patience (Section~3.3), loss function and $\lambda$ (Section~3.2), and architecture details (Section~3.2, Appendix~B) are all specified.

\item {\bf Experiment statistical significance}
    \item[] Question: Does the paper report error bars suitably and correctly defined or other appropriate information about the statistical significance of the experiments?
    \item[] Answer: \answerYes{}
    \item[] Justification: Per-gene $r$ values are reported individually for all 5 held-out genes (Table~1). CTCF enrichment is validated with permutation testing ($P < 0.001$, $n = 1{,}000$). Hi-C validation uses Wilcoxon signed-rank test ($P = 0.020$) and permutation test ($P = 0.088$, $n = 10{,}000$).

\item {\bf Experiments compute resources}
    \item[] Question: For each experiment, does the paper provide sufficient information on the computer resources (type of compute workers, memory, time of execution) needed to reproduce the experiments?
    \item[] Answer: \answerYes{}
    \item[] Justification: Single NVIDIA A100 GPU. CDT-III fine-tuning: $\sim$3 hours; CDT-II pre-training (weight transfer source): $\sim$48 hours (Section~3.3).

\item {\bf Code of ethics}
    \item[] Question: Does the research conducted in the paper conform, in every respect, with the NeurIPS Code of Ethics \url{https://neurips.cc/public/EthicsGuidelines}?
    \item[] Answer: \answerYes{}
    \item[] Justification: The research uses publicly available cell line data (no human subjects), and in silico predictions are explicitly cautioned as not being clinical recommendations (Section~6, Broader impact).

\item {\bf Broader impacts}
    \item[] Question: Does the paper discuss both potential positive societal impacts and negative societal impacts of the work performed?
    \item[] Answer: \answerYes{}
    \item[] Justification: Section~6 (Broader impact) discusses the potential for early drug safety screening and cautions that predictions require experimental validation and should not be used as definitive safety assessments.

\item {\bf Safeguards}
    \item[] Question: Does the paper describe safeguards that have been put in place for responsible release of data or models that have a high risk for misuse (e.g., pre-trained language models, image generators, or scraped datasets)?
    \item[] Answer: \answerNA{}
    \item[] Justification: The model predicts perturbation responses in a single cell line and poses no direct risk for misuse.

\item {\bf Licenses for existing assets}
    \item[] Question: Are the creators or original owners of assets (e.g., code, data, models), used in the paper, properly credited and are the license and terms of use explicitly mentioned and properly respected?
    \item[] Answer: \answerYes{}
    \item[] Justification: STING-seq data (GEO: GSE171452), ENCODE ChIP-seq data, and 4DN Hi-C data are all publicly available and properly cited.

\item {\bf New assets}
    \item[] Question: Are new assets introduced in the paper well documented and is the documentation provided alongside the assets?
    \item[] Answer: \answerNA{}
    \item[] Justification: No new datasets are released. Code will be released upon publication with documentation.

\item {\bf Crowdsourcing and research with human subjects}
    \item[] Question: For crowdsourcing experiments and research with human subjects, does the paper include the full text of instructions given to participants and screenshots, if applicable, as well as details about compensation (if any)?
    \item[] Answer: \answerNA{}
    \item[] Justification: No crowdsourcing or human subjects research (cell line data only).

\item {\bf Institutional review board (IRB) approvals or equivalent for research with human subjects}
    \item[] Question: Does the paper describe potential risks incurred by study participants, whether such risks were disclosed to the subjects, and whether Institutional Review Board (IRB) approvals (or an equivalent approval/review based on the requirements of your country or institution) were obtained?
    \item[] Answer: \answerNA{}
    \item[] Justification: No human subjects research (cell line data only).

\item {\bf Declaration of LLM usage}
    \item[] Question: Does the paper describe the usage of LLMs if it is an important, original, or non-standard component of the core methods in this research? Note that if the LLM is used only for writing, editing, or formatting purposes and does \emph{not} impact the core methodology, scientific rigor, or originality of the research, declaration is not required.
    \item[] Answer: \answerNA{}
    \item[] Justification: LLMs (Claude, Anthropic) were used only for manuscript editing, not as a component of the core methodology. This is disclosed in the Acknowledgments.

\end{enumerate}

\end{document}